\documentclass{article}

\usepackage{times}
\usepackage{graphicx} 
\usepackage{subfig}
\usepackage{natbib}

\usepackage{algorithm}
\usepackage{algorithmic}

\usepackage{hyperref}

\usepackage[accepted]{icml2017}
\usepackage{tablefootnote}
\usepackage{latexsym}
\usepackage{amsmath,amssymb,enumerate,bbm,color,alltt}
\usepackage{url}
\usepackage{graphicx}
\usepackage{float}
\usepackage{stfloats}
\usepackage{epstopdf}
\usepackage{longtable}
\usepackage{tabularx}
\usepackage{multirow}
\usepackage{booktabs}
\usepackage{caption}
\usepackage[titletoc]{appendix}

\icmltitlerunning{Discovering Discrete Latent Topics with Neural Variational Inference}

\begin{document}
\twocolumn[
\icmltitle{Discovering Discrete Latent Topics with Neural Variational Inference}



\icmlsetsymbol{equal}{*}

\begin{icmlauthorlist}
\icmlauthor{Yishu Miao}{ox}
\icmlauthor{Edward Grefenstette}{dm}
\icmlauthor{Phil Blunsom}{ox,dm}
\end{icmlauthorlist}

\icmlaffiliation{ox}{University of Oxford, Oxford, United Kingdom}
\icmlaffiliation{dm}{DeepMind, London, United Kingdom}

\icmlcorrespondingauthor{Yishu Miao}{yishu.miao@cs.ox.ac.uk}

\icmlkeywords{deep learning, variational autoencoder, topic models, bayesian nonparameterics, machine learning, ICML}

\vskip 0.3in
]



\printAffiliationsAndNotice{}  

\begin{abstract}
Topic models have been widely explored as probabilistic generative models of documents.
Traditional inference methods have sought closed-form derivations for updating the models, however as the expressiveness of these models grows, so does the difficulty of performing fast and accurate inference over their parameters.
This paper presents alternative neural approaches to topic modelling by providing parameterisable distributions over topics which permit training by backpropagation in the framework of neural variational inference.
In addition, with the help of a stick-breaking construction, we propose a
  recurrent network that is able to discover a notionally unbounded number of
  topics, analogous to Bayesian non-parametric topic models. Experimental results on the MXM Song Lyrics, 20NewsGroups and Reuters News datasets demonstrate the effectiveness and efficiency of these neural topic models.

\end{abstract}

\section{Introduction}
Probabilistic models for inducing latent topics from documents are one of the
great success stories of unsupervised learning.
Starting with latent semantic analysis (LSA \cite{landauer1998introduction}),
models for uncovering the underlying semantic structure of a document collection
have been widely applied in data mining, text processing and information retrieval.
Probabilistic topic models (e.g.~PLSA \cite{hofmann1999probabilistic}, LDA
\cite{blei2003latent} and HDPs \cite{teh2006hierarchical}) provide a robust,
scalable, and theoretically sound foundation for document modelling by introducing
latent variables for each token to topic assignment.

For the traditional Dirichlet-Multinomial topic model, efficient
inference is available by exploiting conjugacy with either Monte Carlo or
Variational techniques \citep{jordan1999introduction,attias2000variational,beal2003variational}).
However, as topic models have grown more expressive, in order to capture topic
dependencies or exploit conditional information, inference methods have become
increasingly complex. This is especially apparent for non-conjugate models \cite{carlin1991inference, blei2007correlated, wang2013variational}.

Deep neural networks are excellent function approximators and have shown great
potential for learning complicated non-linear distributions for unsupervised models.
Neural variational inference
\citep{kingma2013auto,rezende2014stochastic,mnih2014neural} approximates the
posterior of a generative model with a variational distribution parameterised by
a neural network.
This allows both the generative model and the variational network to be jointly
trained with backpropagation.
For models with continuous latent variables associated with particular
distributions, such as Gaussians, there exist reparameterisations
\cite{kingma2013auto,rezende2014stochastic} of the distribution permitting
unbiased and low-variance estimates of the gradients with respect to the parameters of the inference network.
For models with discrete latent variables, Monte-Carlo estimates of the
gradient must be employed.
Recently, algorithms such as REINFORCE have been used effectively to decrease variance and improve learning \cite{mnih2014neural,mnih2014recurrent}.

In this work we propose and evaluate a range of topic models parameterised with
neural networks and trained with variational inference.
We introduce three different neural structures for constructing topic
distributions: the Gaussian Softmax distribution (GSM), the Gaussian Stick Breaking
distribution (GSB), and the Recurrent Stick Breaking process (RSB), all of which
are conditioned on a draw from a multivariate Gaussian distribution.
The Gaussian Softmax topic model constructs a finite topic distribution with a
softmax function applied to the projection of the Gaussian random vector.
The Gaussian Stick Breaking model also constructs a discrete distribution
from the Gaussian draw, but this time employing a stick breaking construction to provide
a bias towards sparse topic distributions.
Finally, the Recurrent Stick Breaking process employs a recurrent neural
network, again conditioned on the Gaussian draw, to progressively break the
stick, yielding a neural analog of a Dirichlet Process topic model \cite{teh2006hierarchical}.

Our neural topic models combine the merits of both neural networks and traditional probabilistic topic models.
They can be trained efficiently by backpropagation, scaled to large data sets,
and easily conditioned on any available contextual information.
Further, as probabilistic graphical models, they are interpretable and explicitly
represent the dependencies amongst the random variables.
Previous neural document models, such as the neural variational document model (NVDM) \cite{miao2015neural}, belief networks document model \cite{mnih2014neural}, neural auto-regressive document model
\cite{larochelle2012neural} and replicated softmax \cite{hinton2009replicated},
have not explicitly modelled latent topics.
Through evaluations on a range of data sets we compare our models with previously proposed neural document models and traditional probabilistic topic models, demonstrating their robustness and effectiveness.

\section{Parameterising Topic Distributions} \label{sec:cr}
In probabilistic topic models, such as LDA \cite{blei2003latent}, we use the latent variables $\theta_d$ and $z_n$ for the topic proportion of document $d$, and the topic assignment for the observed word $w_n$, respectively.
In order to facilitate efficient inference, the Dirichlet distribution (or Dirichlet process \cite{teh2006hierarchical}) is employed as the prior to generate the parameters of the multinomial distribution $\theta_d$ for each document.
The use of a conjugate prior allows the tractable computation of the posterior
distribution over the latent variables' values.
While alternatives have been explored, such as log-normal topic distributions \cite{blei2006dynamic,blei2007correlated}, extra approximation (e.g.~the Laplace approximation \cite{wang2013variational}) is required for closed form derivations.
The generative process of LDA is:
\begin{eqnarray*}
  & \theta_d \sim \text{Dir}(\alpha_0), & \text{for } d \in D \\
  & z_n \sim \text{Multi}(\theta_d), & \text{for } n \in [1,N_d] \\
  & w_n \sim \text{Multi}(\beta_{z_n}), & \text{for } n \in [1,N_d]
\end{eqnarray*}
where $\beta_{z_n}$ represents the topic distribution over words given topic assignment $z_n$ and $N_d$ is the number of tokens in document $d$.
$\beta_{z_n}$ can be drawn from another Dirichlet distribution, but here we
consider it a model parameter.
$\alpha_0$ is the hyper-parameter of the Dirichlet prior and $N_d$ is the total number of words in document $d$.
The marginal likelihood for a document in collection $D$ is:
\begin{eqnarray}
  p(d|\alpha_0, \beta)\!\! = \!\!\! \int_{\theta} \! p(\theta|\alpha_0) \! \prod_n \! \sum_{z_n} p(w_n|\beta_{z_n})p(z_n|\theta) d\theta .
  \label{eq:lh1}
\end{eqnarray}
If we employ mean-field variational inference, the updates for the variational parameters $q(\theta)$ and $q(z_n)$ can be directly derived in closed form.

In contrast, our proposed models introduce a neural network to parameterise the
multinomial topic distribution. The generative process is:
\begin{eqnarray*}
  & \theta_d \sim \text{G}(\mu_0,\sigma_0^2), & \text{for } d \in D \\
  & z_n \sim \text{Multi}(\theta_d), & \text{for } n \in [1,N_d] \\
  & w_n \sim \text{Multi}(\beta_{z_n}), & \text{for } n \in [1,N_d]
\end{eqnarray*}
where $G(\mu_0,\sigma_0^2)$ is composed of a neural network $\theta = g(x)$
conditioned on a isotropic Gaussian $x \sim
N(\mu_0,\sigma_0^2)$\footnote{Throughout this presentation we employ diagonal
Gaussian distributions. As such we use $N(\mu,\sigma^2)$ to represent the
Gaussian distributions, where $\sigma^2$ is the diagonal of the covariance matrix.}.
The marginal likelihood is:
\begin{eqnarray}
  p(d|\mu_0,\sigma_0, \beta) &=& \int_{\theta}  p(\theta|\mu_0,\sigma_0^2) \label{eq:lh2} \\
  &&\prod\nolimits_n  \sum\nolimits_{z_n} p(w_n|\beta_{z_n})p(z_n|\theta) d\theta \nonumber.
\end{eqnarray}
Compared to Equation (\ref{eq:lh1}), here we parameterise the latent variable
$\theta$ by a neural network conditioned on a draw from a Gaussian distribution.
To carry out neural variational inference \cite{miao2015neural}, we construct an inference network $q(\theta|\mu(d), \sigma(d))$ to approximate the posterior $p(\theta|d)$, where $\mu(d)$ and $\sigma(d)$ are functions of $d$ that are implemented by multilayer perceptrons (MLP).
By using a Gaussian prior distribution, we are able to employ the re-parameterisation trick \cite{kingma2013auto} to build an unbiased and low-variance gradient estimator for the variational distribution.
Without conjugacy, the updates of the parameters can still be derived directly and easily from the variational lower bound.
We defer discussion of the inference process until the next section. Here we introduce several alternative neural networks for $g(x)$ which transform a Gaussian sample $x$ into the topic proportions $\theta$.

\subsection{The Gaussian Softmax Construction}
In deep learning, an energy-based function is generally used to construct probability distributions \cite{lecun2006tutorial}.
Here we pass a Gaussian random vector through a softmax function to parameterise the multinomial document topic distributions.
Thus $\theta \sim G_\text{GSM}(\mu_0,\sigma_0^2)$ is defined as:
\begin{eqnarray*}
   x \sim \mathcal{N}(\mu_0,\sigma_0^2)  \\
  \theta = \text{softmax}(W_1^T x)
\end{eqnarray*}
where $W_1$ is a linear transformation, and we leave out the bias terms for brevity.
$\mu_0$ and $\sigma_0^2$ are hyper-parameters which we set for a zero
mean and unit variance Gaussian.

\subsection{The Gaussian Stick Breaking Construction}
In Bayesian non-parametrics, the stick breaking process \cite{sethuraman1994constructive} is used as a constructive definition of the Dirichlet process, where sequentially drawn Beta random variables define breaks from a unit stick.
In our case, following \citet{khan2012stick}, we transform the modelling of multinomial probability parameters into the modelling of the logits of binomial probability parameters using Gaussian latent variables.
More specifically, conditioned on a Gaussian sample $x \in \mathbbm{R}^{H}$, the breaking proportions $\eta \in \mathbbm{R}^{K-1}$ are generated by applying the sigmoid function $\eta = \text{sigmoid}(W_2^T x)$
where $W \in \mathbbm{R}^{H \times ï¼ˆK-1ï¼‰}$.
Starting with the first piece of the stick, the probability of the first category is modelled as a break of proportion $\eta_1$, while the length of the remainder of the stick is left for the next break.
Thus each dimension can be deterministically computed by $\theta_{k} = \eta_{k} \prod_{i = 1}^{k - 1} (1 - \eta_i)$ until $k\!\!=\!\!K\!\!-\!\!1$, and the remaining length is taken as the probability of the $K$th category $\theta_K = \prod_{i = 1}^{K - 1} (1 - \eta_i)$.

\begin{figure}[t]
  \centering
  \vspace{-0.8em}
	\includegraphics[width=3.0in]{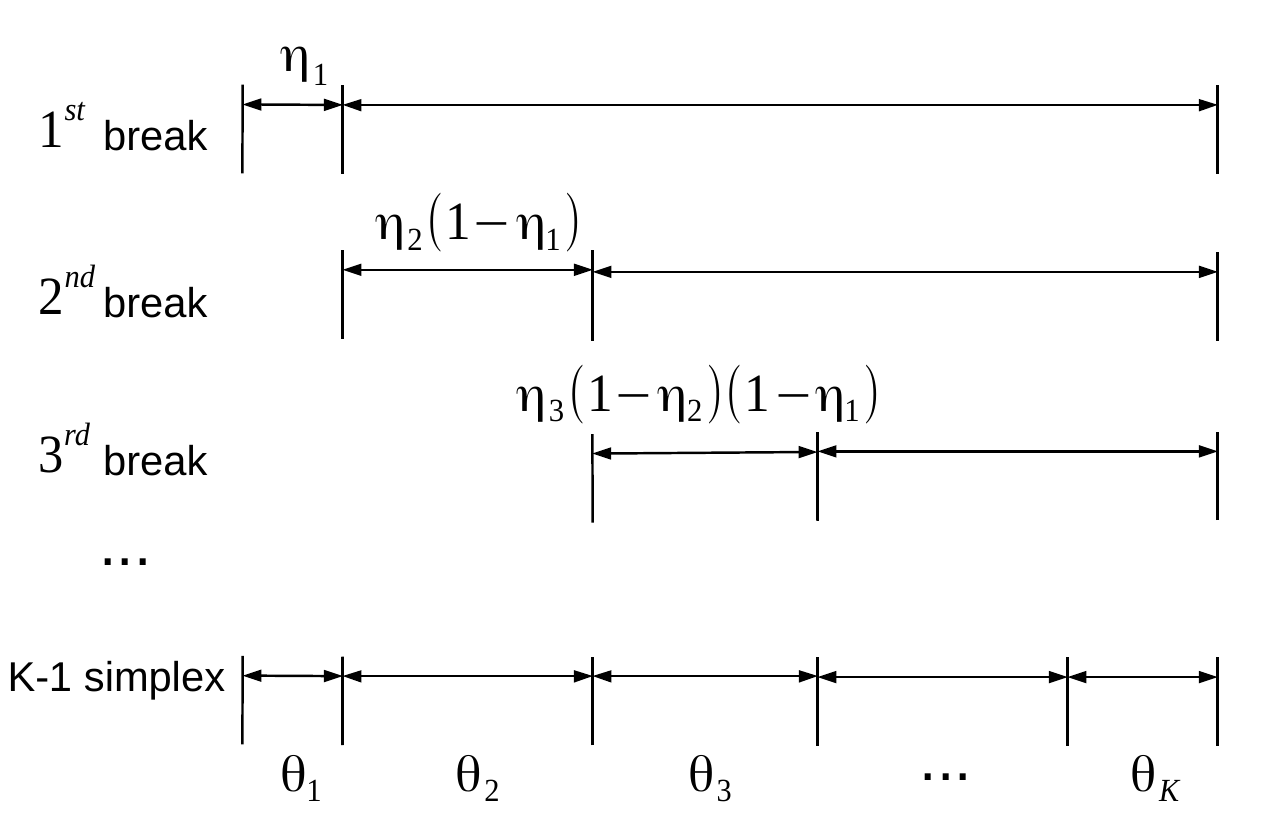}
	\vspace{-1.0em}
  \caption{The Stick Breaking Construction.}
  \label{fig:sb}
  	\vspace{-1.0em}
\end{figure}

For instance assume $K=3$, $\theta$ is generated by 2 breaks where $\theta_1=\eta_1$, $\theta_2=\eta_2(1-\eta_1)$ and the remaining stick $\theta_3=(1-\eta_2)(1-\eta_1)$. If the model proceeds to break the stick for $K=4$, the remaining stick $\theta_3$ is broken into $(\theta_3', \theta_4')$, where $\theta_3'=\eta_3\cdot \theta_3$, $\theta_4'=(1-\eta_3)\cdot \theta_3$ and $\theta_3=\theta_3' + \theta_4'$. Hence, for different values of $K$, it always satisfies $\sum_{k=1}^K\theta_k=1$.
The stick breaking construction $f_\text{SB}(\eta)$ is illustrated in Figure \ref{fig:sb} and the distribution $\theta \sim G_\text{GSB}(\mu_0,\sigma_0^2)$ is defined as:
\begin{eqnarray*}
  x & \sim & \mathcal{N} (\mu_0, \sigma_0^2)\\
  \eta & = & \text{sigmoid}(W_2^T x)\\
  \theta & = & f_\text{SB} (\eta)
\end{eqnarray*}

Although the Gaussian stick breaking construction breaks exchangeability, compared to the stick breaking definition of the Dirichlet process, it does provide a more amenable form for neural variational inference.
More interestingly, this stick breaking construction introduces a non-parametric aspect to neural topic models.

\subsection{The Recurrent Stick Breaking Construction}
\begin{figure}[t]
  \centering
  \vspace{-0.3em}
	\includegraphics[width=2.85in]{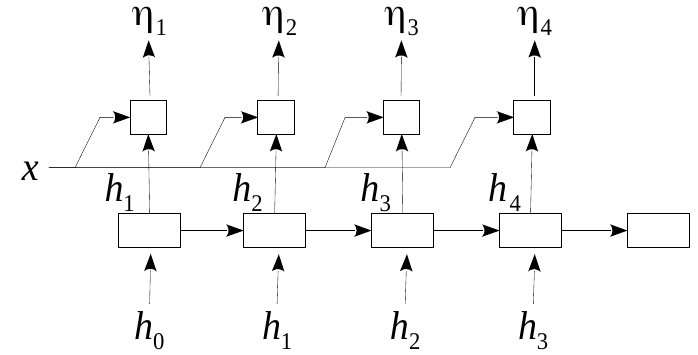}
  \caption{The unrolled Recurrent Neural Network that produces the stick breaking proportions $\eta$.}
  \label{fig:rnn1}
  	\vspace{-0.5em}
\end{figure}

Recurrent Neural Networks (RNN) are commonly used for modelling sequences of inputs in deep learning.
Here we consider the stick breaking construction as a sequential draw from an
RNN, thus capturing an unbounded number of breaks with a finite number of
parameters.
Conditioned on a Gaussian latent variable $x$, the recurrent neural network
$f_\text{SB}(x)$ produces a sequence of binomial logits which are used to break
the stick sequentially. The $f_\text{RNN}(x)$ is decomposed as:
\begin{eqnarray*}
  h_k & = &  \text{RNN}_\text{SB} (h_{k-1}) \\
  \eta_k & = & \text{sigmoid} (h_{k-1}^T x)
\end{eqnarray*}
where $h_k$ is the output of the $k$th state, which we feed into the next state of the RNN$_\text{SB}$ as an input.
Figure \ref{fig:rnn1} shows the recurrent neural network structure. Now $\theta \sim G_\text{RSB}(\mu_0,\sigma_0^2)$ is defined as:
\begin{eqnarray*}
  x & \sim & \mathcal{N} (\mu_0, \sigma_0^2)\\
  \eta & = & f_\text{RNN}(x)\\
  \theta & = & f_\text{SB} (\eta)
\end{eqnarray*}
where $f_\text{SB}(\eta)$ is equivalent to the stick breaking function used in the Gaussian stick breaking construction.
Here, the RNN is able to dynamically produce new logits to break the stick {\em ad infinitum}.
The expressive power of the RNN to model sequences of unbounded length
is still bounded by the parametric model's capacity, but for topic modelling it is adequate to model the countably infinite topics amongst the documents in a truncation-free fashion.

\section{Models}
Given the above described constructions for the topic distributions, in this section we introduce our family of neural topic models and corresponding inference methods.
\subsection{Neural Topic Models}
Assume we have finite number of topics $K$, the topic distribution over words given a topic assignment $z_n$ is $p(w_n|\beta, z_n)=\text{Multi}(\beta_{z_n})$. Here we introduce topic vectors $t \in \mathbbm{R}^{K\times H}$, word vectors $v \in \mathbbm{R}^{V\times H}$ and generate the topic distributions over words by:
\begin{eqnarray*}
   \beta_{k} = \text{softmax}(v \cdot t^T_{k}).
\end{eqnarray*}
Therefore, $\beta \! \in \! \mathbbm{R}^{K\times V}$ is a collection of simplexes achieved by computing the semantic similarity between topics and words.
Following the notation introduced in Section \ref{sec:cr}, the prior distribution is defined as $G(\theta|\mu_0,\sigma_0^2)$ in which $x \sim \mathcal{N}(x|\mu_0,\sigma_0^2)$ and the projection network generates $\theta=g(x)$ for each document.
Here, $g(x)$ can be the Gaussian Softmax $g_\text{GSM}(x)$, Gaussian Stick Breaking $g_\text{GSB}(x)$, or Recurrent Stick Breaking $g_\text{RSB}(x)$ constructions with fixed length RNN$_\text{SB}$.
We derive a variational lower bound for the document log-likelihood according to Equation (\ref{eq:lh2}):
\begin{eqnarray}
   & \mathcal{L}_d = \!\!\! & \mathbbm{E}_{q(\theta|d)}
  \left[ \sum\nolimits_{n = 1}^N \log \sum\nolimits_{z_n} [p(w_{n}|\beta_{z_n}) p(z_n|\theta) ] \right] \nonumber \\
  &  & - D_{KL}
  \left[ q(\theta|d) || p(\theta|\mu_0,\sigma_0^2) \right] \label{eq:lb1}
\end{eqnarray}
where $q(\theta|d)$ is the variational distribution approximating the true posterior $p(\theta|d)$.
Following the framework of neural variational inference \cite{miao2015neural,kingma2013auto,rezende2014stochastic}, we introduce an inference network conditioned on the observed document $d$ to generate the variational parameters $\mu(d)$ and $\sigma(d)$ so that we can estimate the lower bound by sampling $\theta$ from $q(\theta|d)=G(\theta|\mu(d),\sigma^2(d))$.
In practise we reparameterise $\hat{\theta}=\mu(d)+\hat{\epsilon} \cdot \sigma(d)$ with the sample $\hat{\epsilon} \in \mathcal{N}(0,I)$.

\begin{figure}[t]
  \centering
  \vspace{-0.8em}
	\includegraphics[width=3.4in]{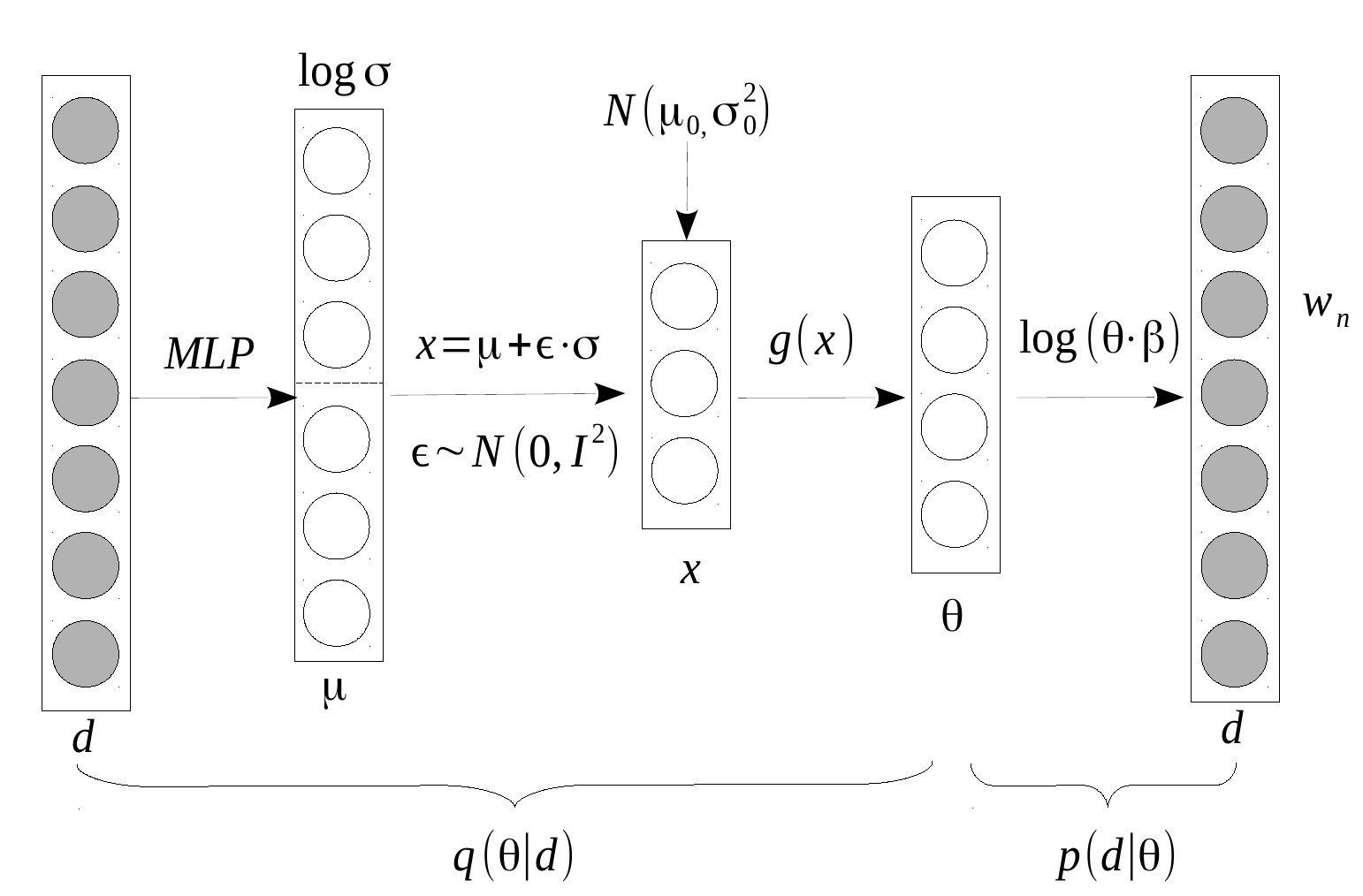}
  \caption{Network structure of the inference model $q(\theta \mid d)$, and of the generative model $p(d \mid \theta)$.}
  \label{fig:ntm}
  	\vspace{-0.5em}
\end{figure}

Since the generative distribution $p(\theta|\mu_0,\sigma_0^2)=p(g(x)|\mu_0,\sigma_0^2)=p(x|\mu_0,\sigma_0^2)$ and the variational distribution $q(\theta|d)=q(g(x)|d)=q(x|\mu(d),\sigma^2(d))$, the KL term in Equation (\ref{eq:lb1}) can be easily integrated as a Gaussian KL-divergence.
Note that, the parameterisation network $g(x)$ and its parameters are shared across all the documents.
In addition, given a sampled $\hat{\theta}$, the latent variable $z_n$ can be integrated out as:
\begin{eqnarray}
 \log p(w_{n}|\beta, \hat{\theta})&=&\log \sum\nolimits_{z_n} \left[p(w_{n}|\beta_{z_n}) p(z_n|\hat{\theta}) \right]\nonumber \\
 &=&\log (\hat{\theta} \cdot \beta)
   \label{eq:int}
\end{eqnarray}
Thus there is no need to introduce another variational approximation for the topic assignment $z$.
The variational lower bound is therefore:
\begin{eqnarray*}
   \mathcal{L}_d \approx \hat{\mathcal{L}}_d =  \!  \sum\nolimits_{n = 1}^N
  \left[  \log p(w_{n}|\beta, \hat{\theta}) \right]  \label{eq:lb2}  - D_{KL}
  \left[ q(x|d) || p(x) \right]  \nonumber
\end{eqnarray*}
We can directly derive the gradients of the generative parameters $\Theta$, including $t$, $v$ and $g(x)$. While for the variational parameters $\Phi$, including $\mu(d)$ and $\sigma(d)$, we use the gradient estimators:
\begin{eqnarray*}
    \nabla_{\mu(d)}\hat{\mathcal{L}}_d &\approx & \nabla_{\hat{\theta}}\hat{\mathcal{L}}_d ,  \\
    \nabla_{\sigma(d)}\hat{\mathcal{L}}_d &\approx & \hat{\epsilon} \cdot \nabla_{\hat{\theta}}\hat{\mathcal{L}}_d .
\end{eqnarray*}
$\Theta$ and $\Phi$ are jointly updated by stochastic gradient back-propagation.
The structure of this variational auto-encoder is illustrated in Figure \ref{fig:ntm}.

\subsection{Recurrent Neural Topic Models}
For the GSM and GSB models the topic vectors $t \in \mathbbm{R}^{K \times H}$ have to be predefined for computing the topic distribution over words $\beta$.
With the RSB construction we can model an unbounded number of topics, however in addition to the RNN$_\text{SB}$ that generates the topic proportions $\theta \in \mathbbm{R^\infty}$ for each document, we must introduce another neural network RNN$_\text{Topic}$ to produce the topics $t \in \mathbbm{R}^{\infty \times H}$ dynamically, so as to avoid the need to truncate the variational inference.

\begin{figure}[t]
  \centering
  \vspace{-0.8em}
	\includegraphics[width=3.0in]{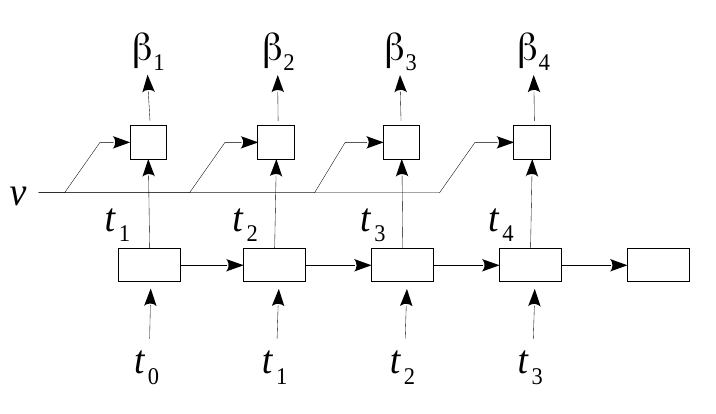}
	\vspace{-1.5em}
  \caption{The unrolled Recurrent Neural Network that produces the topic-word distributions $\beta$.}
  \label{fig:rnn2}
  	\vspace{-0.5em}
\end{figure}

For comparison, in finite neural topic models we have topic vectors $t \in \mathbbm{R}^{K\times H}$, while in unbounded neural topic models the topics $t \in \mathbbm{R}^{\infty \times H }$ are dynamically generated by RNN$_\text{Topic}$ and the order of the topics corresponds to the order of the states in RNN$_\text{SB}$. The generation of $\beta$ follows:
\begin{eqnarray*}
  t_k & = &  \text{RNN}_\text{Topic} (t_{k-1}) , \\
  \beta_k & = & \text{softmax} (v \cdot t_{k}^T) ,
\end{eqnarray*}
where $v \in \mathbbm{R}^{V \times H}$ represents the word vectors, $t_k$ is the $k$th topic generated by RNN$_\text{Topic}$ and $k<\infty$.
Figure \ref{fig:rnn2} illustrates the neural structure of RNN$_\text{Topic}$.

For the unbounded topic models we introduce a truncation-free neural variational inference method which enables the model to dynamically decide the number of active topics.
Assume the current active number of topics is $i$, RNN$_\text{Topic}$ generates $t^i \in \mathbbm{R}^{i \times H }$ by an $i-1$ step stick-breaking process (the logit for the $n$th topic is the remaining stick after $i-1$ breaks).
The variational lower bound for a document $d$ is:
\begin{eqnarray*}
   \mathcal{L}^i_d \approx \sum\nolimits_{n = 1}^N
  \left[  \log p(w_{n}|\beta^i, \hat{\theta}^i) \right]  - D_{KL}
  \left[ q(x|d) || p(x) \right] ,
\end{eqnarray*}
where $\hat{\theta}^i$ corresponds to the topic distribution over words $\beta^t$.
In order to dynamically increase the number of topics, the model proposes the $i$th break on the stick to split the $(i+1)$th topic.
In this case, RNN$_\text{Topic}$ proceeds to the next state and generates topic $t^{(i+1)}$ for $\beta^{(i+1)}$ and the RNN$_\text{SB}$ generates $\hat{\theta}^{(i+1)}$ by an extra break of the stick.
Firstly, we compute the likelihood increase brought by topic $i$ across the documents $D$:
\begin{eqnarray*}
   \mathcal{I} = \sum\nolimits_{d}^D \left[\mathcal{L}^{i}_d - \mathcal{L}^{i-1}_d \right] / \sum\nolimits_{d}^D [\mathcal{L}^{i}_d]
\end{eqnarray*}
Then, we employ an acceptance hyper-parameter $\gamma$ to decide whether to generate a new topic. If $\mathcal{I}>\gamma$, the previous proposed new topic (the $i$th topic) contributes to the generation of words and we increase the active number of topics $i$ by 1, otherwise we keep the current $i$ unchanged.
Thus $\gamma$ controls the rate at which the model generates new topics. In practise, the increase of the lower bound is computed over mini-batches so that the model is able to generate new topics before the current epoch is finished. The details of the algorithm are described in Algorithm \ref{al:rntm}.

\subsection{Topic vs.\ Document Models} \label{sec:itd}

In most topic models, documents are modelled by a mixture of topics, and each word is associated with a single topic latent variable, e.g. LDA and GSM. 
However, the NVDM \cite{miao2015neural} is implemented as a VAE \cite{kingma2013auto} without modelling topics explicitly.
The major difference is that NVDM employs a softmax decoder (Equation \ref{eq:sm}) to generate all of the words of a document conditioned on the document representation $\hat{\theta}$:
\begin{eqnarray}
 \log p(w_{n}|\beta, \hat{\theta})=\log \text{softmax}(\hat{\theta} \cdot \beta).
   \label{eq:sm}
\end{eqnarray}
where both $\hat{\theta}$ and $\beta$ are unnormalised.
Hence, it breaks the topic model assumption that each document consists of a mixture of topics. 
Although the latent variables can still be interpreted as topics,
these topics are not modelled explicitly since there is no actual topic distribution over words.
\citet{srivastavaneural} interprets the above decoder as a weighted product of experts topic model, here however, we refer to such models that do not directly assign topics to words as document models instead of topic models.
We can also convert our neural topic models to neural document models by replacing the mixture decoder (Equation \ref{eq:int}) with the softmax decoder (Equation \ref{eq:sm}).
For example in the GSM construction, if we remove the softmax function over $\theta$, and directly apply Equation \ref{eq:sm} to generate the words, it reduces to a variant of the NVDM (GSM applies topic and word vectors to compute $\beta$, while NVDM directly models a projection $\beta$ from the latent variables to words).

\begin{algorithm}[t]
\footnotesize
\caption{Unbounded Recurrent Neural Topic Model}
\begin{algorithmic}[1]
	\item Initialise $\Theta$ and $\Phi$;  Set active topic number $i$
	\REPEAT
	\FOR{$s \in \text{minibatches } S $}
		\FOR{$ k \in  [1, i] $}
		\STATE{ Compute topic vector $t_k = \text{RNN}_\text{Topic}(t_{k-1})$ }
		\STATE{ Compute topic distribution  $\beta_k = \text{softmax}(v \cdot t_k^T)$}
		\ENDFOR
    \FOR{$d\in  D_s$}
      \STATE{ Sample topic proportion  $\hat{\theta}  \sim  G_\text{RSB} (\theta | \mu(d), \sigma^2(d))$}
      \FOR{$w\in \text{document } d$}
      		 \STATE{ Compute log-likelihood  $\log p(w|\hat{\theta}, \beta) $}
      \ENDFOR
      \STATE{ Compute lowerbound $\mathcal{L}_d^{i-1}$ and $\mathcal{L}_d^{i}$ }
      \STATE{ Compute gradients $\nabla_{\Theta,\Phi} \mathcal{L}_d^{i}$ and update}
    \ENDFOR
    \STATE{ Compute likelihood increase $\mathcal{I}$}
    \IF {$\mathcal{I}>\gamma$}
         \STATE{ Increase active topic number $i=i+1$}
    \ENDIF
    \ENDFOR
    \UNTIL{Convergence}
\end{algorithmic}
\label{al:rntm}
\vspace{-0.5em}
\end{algorithm}

\section{Related Work}
Topic models have been extensively studied for a variety of applications in document modelling and information retrieval.
Beyond LDA, significant extensions have sought to capture topic correlations \cite{blei2007correlated}, model temporal dependencies \cite{blei2006dynamic} and discover an unbounded number of topics \cite{teh2006hierarchical}.
Topic models have been extended to capture extra context information such as time \cite{wang2006topics}, authorship \cite{rosen2004author}, and class labels \cite{mcauliffe2008supervised}. 
Such extensions often require carefully tailored graphical models, and associated inference algorithms, to capture the desired context.
Neural models provide a more generic and extendable option and a number of works
have sought to leverage these, such as the Replicated Softmax \citep{hinton2009replicated}, the Auto-Regressive Document Model \citep{larochelle2012neural}, Sigmoid Belief Document Model \cite{mnih2014neural}, Variational Auto-Encoder Document Model (NVDM) \cite{miao2015neural} and TopicRNN Model \cite{dieng2016topicrnn}.
However, these neural works do not explicitly capture topic assignments.

The recent work of \citet{srivastavaneural} also employs neural variational inference to train topic models and is closely related to our work.
Their model follows the original LDA formulation in keeping the Dirichlet-Multinomial parameterisation and applies a Laplace approximation to allow gradient to back-propagate to the variational distribution.
In contrast, our models directly parameterise the multinomial distribution with neural networks and jointly learn the model and variational parameters during inference.
\citet{nalisnick2016deep} proposes a reparameterisation approach for continuous latent variables with Beta prior, which enables neural variational inference for Dirichlet process.
However, Taylor expansion is required to approximate the KL Divergence while having  multiple draws from the Kumaraswamy variational distribution. In our case, we can easily apply the Gaussian reparametersation trick with only one draw from the Gaussian distribution.

\begin{table}[t]
\center
\addtolength{\tabcolsep}{-4.5pt}
\begin{tabular}{lcccccc}
\toprule[1.5pt]
	\multirow{2}{3cm}{\textbf{Finite Topic Model}} & \multicolumn{2}{c}{MXM} & \multicolumn{2}{c}{20News} & \multicolumn{2}{c}{RCV1} \\
	\cmidrule{2-3}\cmidrule{4-5} \cmidrule{6-7}
	&	50 &	200  &	50 &	200 &	50 &	200 \\
	\hline \\[-2ex]
	\hspace{0.2cm}GSM & \textbf{306} & \textbf{272} &	\textbf{822}	& 830 &	\textbf{717} & \textbf{602}\\
	\hspace{0.2cm}GSB & 309 & 296&	838	& 826 &	788 & 634 \\
	\hspace{0.2cm}RSB & 311 & 297&	835	& \textbf{822} &	750 & 628\\
	\hline \\[-2ex]
	\hspace{0.2cm}OnlineLDA & 312 & 342 &	893 &	1015 &	1062 &	1058 \\
	\multicolumn{2}{l}{\cite{hoffman2010online}} & \\
	\hspace{0.2cm}NVLDA & 330 & 357& 1073 &	993	& 791 & 797\\
	\multicolumn{2}{l}{\cite{srivastavaneural}} & \\
	\midrule[1.2pt]
	\textbf{Unbounded Topic Model}  & \multicolumn{2}{c}{MXM} & \multicolumn{2}{c}{20News} & \multicolumn{2}{c}{RCV1} \\
	\hline \\[-2ex]
	\hspace{0.2cm}RSB-TF & \multicolumn{2}{c}{\textbf{303}} & \multicolumn{2}{c}{\textbf{825}} & \multicolumn{2}{c}{\textbf{622}}\\
	\hspace{0.2cm}HDP \cite{wang2011online} & \multicolumn{2}{c}{370} &	 \multicolumn{2}{c}{937} & \multicolumn{2}{c}{918}\\
	[-0.5ex]
\bottomrule[1.5pt]
\end{tabular}
\caption{Perplexities of the topic models on the test datasets.
The upper section of the table lists the results for finite neural topic models, with 50 or 200 topics, on the MXM, 20NewsGroups and RCV1 datasets.
We compare our neural topic models with the Gaussian Softmax (GSM), Gaussian Stick Breaking (GSB) and Recurrent Stick Breaking (RSB) constructions to the online variational LDA (onlineLDA) \cite{hoffman2010online} and neural variational inference LDA (NVLDA) \cite{srivastavaneural} models.
The lower section shows the results for the unbounded topic models, including our truncation-free RSB (RSB-TF) and the online HDP topic model \cite{wang2011online}.
}
\label{tb:ntm}
\vspace{-1em}
\end{table}
\begin{table}[t]
\center
\addtolength{\tabcolsep}{-4.6pt}
\begin{tabular}{lcccccc}
\toprule[1.5pt]
	\multirow{2}{3.6cm}{\textbf{Finite Document Model}} & \multicolumn{2}{c}{MXM} &\multicolumn{2}{c}{20News} & \multicolumn{2}{c}{RCV1} \\
	\cmidrule{2-3}\cmidrule{4-5} \cmidrule{6-7}
	 &	50 &	200&	50 &	200 &	50 &	200 \\
	\hline \\[-2ex]
	\hspace{0.2cm}GSM & \textbf{270} & \textbf{267} &	787	& 829 &	\textbf{653} & \textbf{521}\\
	\hspace{0.2cm}GSB & 285 & 275&	816	& 815 &	712 & 544\\
	\hspace{0.2cm}RSB & 286 & 283&	\textbf{785}	& \textbf{792} &	662 & 534\\
	\hline \\[-2ex]
	\hspace{0.2cm}NVDM & 345 & 345 &	837 & 873	 &	717 &	588 \\
	\multicolumn{2}{l}{\cite{miao2015neural}} & \\
	\hspace{0.2cm}ProdLDA &319 & 326 &	1009 &	989 & 780 & 788	\\
	\multicolumn{2}{l}{\cite{srivastavaneural}}& \\
	\midrule[1.2pt]
	\textbf{Unbounded Document Model}  & \multicolumn{2}{c}{MXM} & \multicolumn{2}{c}{20News} & \multicolumn{2}{c}{RCV1} \\
	\hline \\[-2ex]
	\hspace{0.2cm}RSB-TF & \multicolumn{2}{c}{285} & \multicolumn{2}{c}{788} & \multicolumn{2}{c}{532}\\
\bottomrule[1.5pt]
\end{tabular}
\caption{Perplexities of document models on the test datasets.
  The table compares the results for a fixed dimension latent variable, 50 or 200, achieved by our neural document models to Product of Experts LDA (prodLDA) \cite{srivastavaneural} and the Neural Variational Document Model (NVDM) \cite{miao2015neural}.
}
\vspace{-1em}
\label{tb:ndm}
\end{table}

\section{Experiments}

We perform an experimental evaluation employing three datasets: \textit{MXM}\footnote{http://labrosa.ee.columbia.edu/millionsong/musixmatch \cite{Bertin-Mahieux2011}} song lyrics, \textit{20NewsGroups}\footnote{http://qwone.com/~jason/20Newsgroups} and Reuters \textit{RCV1-v2}\footnote{http://trec.nist.gov/data/reuters/reuters.html} news.
\textit{MXM} is the official lyrics collection of the Million Song Dataset with 210,519 training and 27,143 testing datapoints respectively.
The \textit{20NewsGroups} corpus is divided into 11,314 training and 7,531 testing documents, while the \textit{RCV1-v2} corpus is a larger collection with 794,414 training and 10,000 test cases from Reuters newswire stories.
We employ the original 5,000 vocabulary provided for \textit{MXM}, while the other two datasets are processed by stemming, filtering stopwords and taking the most frequent 2,000\footnote{We use the vocabulary provided by \citet{srivastavaneural} for direct comparison.} and 10,000 words as the vocabularies.

The hidden dimension of the MLP for constructing $q(\theta|d)$ is 256 for all the neural topic models and the benchmarks that apply neural variational inference (e.g.~NVDM, proLDA, NVLDA), and 0.8 dropout is applied on the output of the MLP before parameterising the diagonal Gaussian distribution.
Grid search is carried out on learning rate and batch size for achieving the held-out perplexity.
For the recurrent stick breaking construction we use a one layer LSTM cell (256 hidden units) for constructing the recurrent neural network.
For the finite topic models we set the maximum number of topics $K$ as 50 and 200.
The models are trained by Adam \citep{DBLP:journals/corr/KingmaB14} and only one sample is used for neural variational inference.
We follow the optimisation strategy of \citet{miao2015neural} by alternately updating the model parameters and the inference network. To alleviate the redundant topics issue, we also apply topic diversity regularisation \cite{xie2015diversifying} while carrying out neural variational inference (Appendix \ref{a:td}).

\subsection{Evaluation}
We use Perplexity as the main metric for assessing the generalisation ability of our generative models.
Here we use the variational lower bound to estimate the document perplexity: $\text{exp}(-\frac{1}{D}\sum_d^{D} \frac{1}{N_d} \log p(d))$ following \citet{miao2015neural}.
Table \ref{tb:ntm} presents the test document perplexities of the topic models on the three datasets.
Amongst the finite topic models, the Gaussian softmax construction (GSM) achieves the lowest perplexity in most cases, while all of the GSM, GSB and RSB models are significantly better than the benchmark LDA and NVLDA models.
Amongst our selection of unbounded topic models, we compare our truncation-free RSB model, which applies an RNN to dynamically increase the active topics ($\gamma$ is empirically set as $5e^{-5}$), with the traditional non-parametric HDP topic model \cite{teh2006hierarchical}.
Here we see that the recurrent neural topic model performs significantly better than the HDP topic model on perplexity.

Next we evaluate our neural network parameterisations as document models with the implicit topic distribution introduced in Section \ref{sec:itd}.
Table \ref{tb:ndm} compares the proposed neural document models with the benchmarks.
According to our experimental results, the generalisation abilities of the GSM, GSB and RSB models are all improved by switching to an implicit topic distribution, and their performance is also significantly better than the NVDM and ProdLDA.
We hypothesise that this effect is due to the models not needing to infer the topic-word assignments, which makes optimisation much easier.
Interestingly, the RSB model performs better than the GSM and GSB on \textit{20NewsGroups} in both the 50 and 200 topic settings.
This is possibly due to the fact that GSM and GSB apply linear transformations $W_1$ and $W_2$ to generate the hidden variable $\theta$ and breaking proportions $\eta$ from a Gaussian draw, while the RSB applies recurrent neural networks to produce $\eta$ in a sequence which induces dependencies in $\eta$ and helps escape local minima.
It is worth noting that the recurrent neural network uses more parameters than the other two models.
As mentioned in Section \ref{sec:itd}, GSM is a variant of NVDM that applies topic and word vectors to construct the topic distribution over words instead of directly modelling a multinomial distribution by a softmax function, which further simplifies optimisation.
If it is not necessary to model the explicit topic distribution over words, using an implicit topic distribution may lead to better generalisation.

\begin{figure}[t]
  \centering
  \vspace{-0.2em}
  \subfloat[GSM Model]{\includegraphics[width=0.38\textwidth]{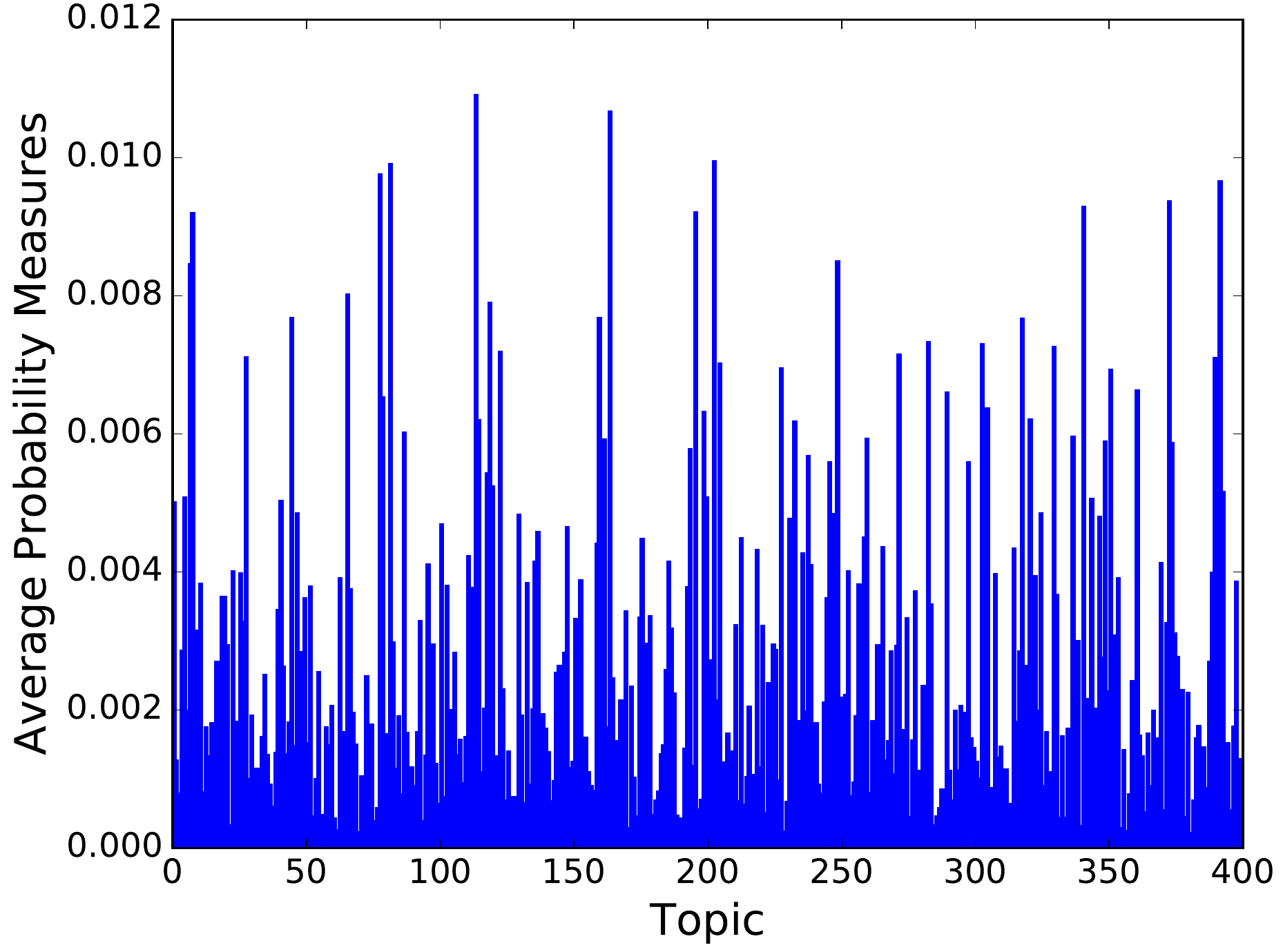}\label{fig:pmgsm}}
  \vspace{-1em}
  \hfill
  \subfloat[GSB Model]{\includegraphics[width=0.38\textwidth]{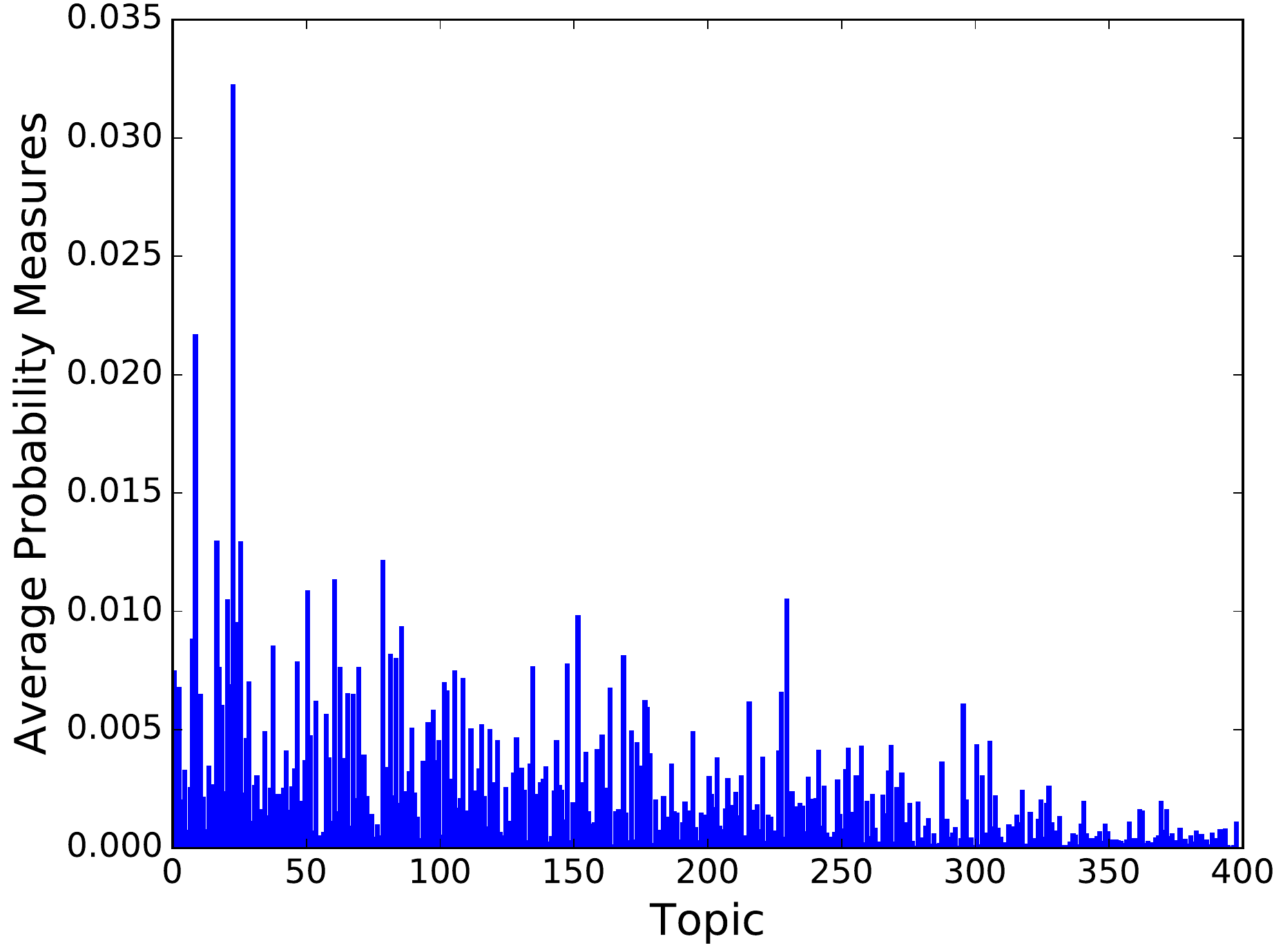}\label{fig:pmgsb}}
  \caption{Corpus level topic probability distributions.}
  \vspace{-1.0em}
\label{fig:pm}
\end{figure}

\begin{figure}[t]
  \centering
	\includegraphics[width=3.0in]{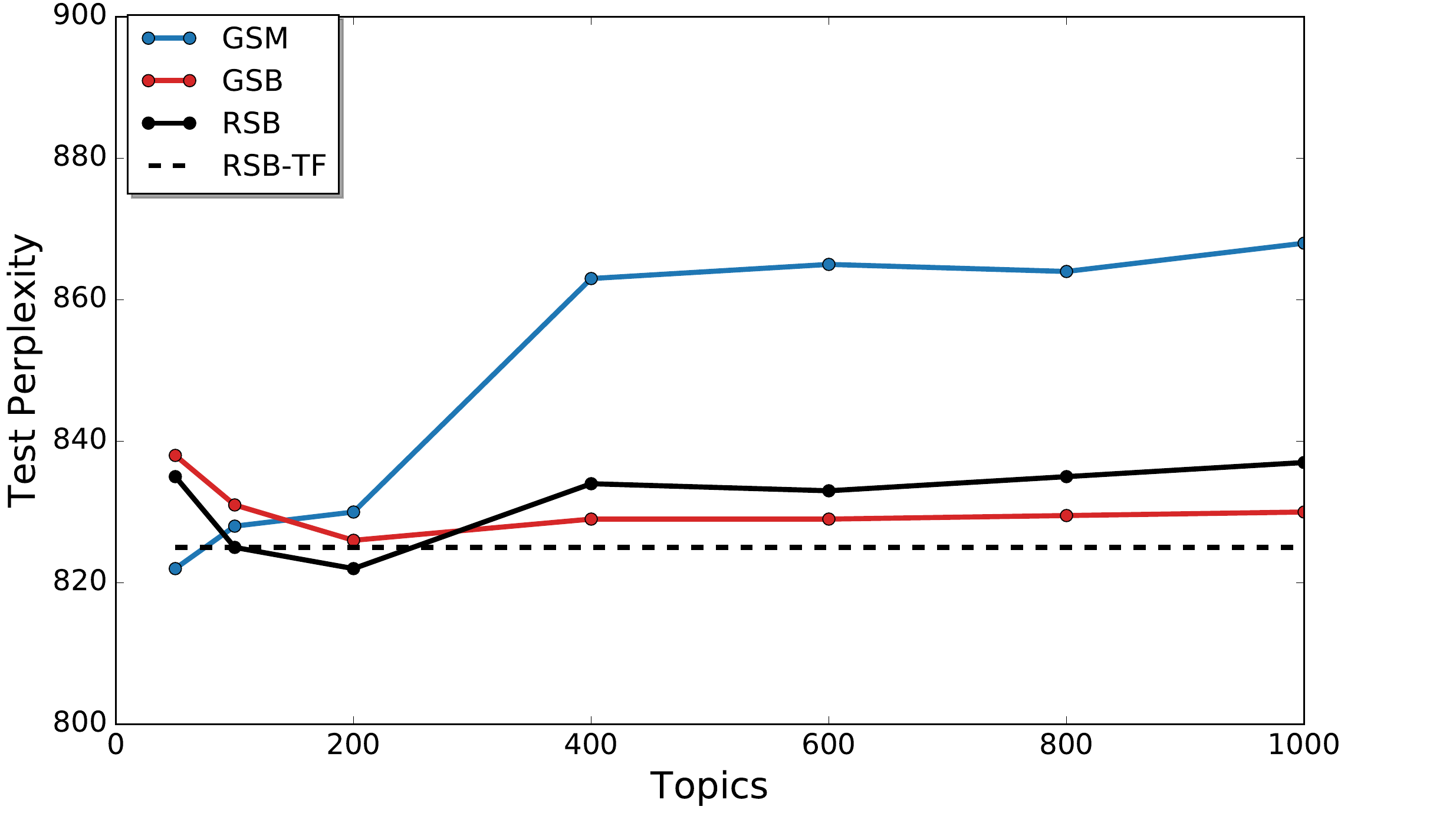}
	\vspace{-1em}
  \caption{Test perplexities of the neural topic models with a varying maximum number of topics on the 20NewsGroups dataset. The truncation-free RSB (RSB-TF) dynamically increases the active topics, we use a dashed line to represent its test perplexity for reference in the figure.}
  \label{fig:np}
  	\vspace{-0.5em}
\end{figure}

To further demonstrate the effectiveness of the stick-breaking construction, Figure \ref{fig:pm} presents the average probability of each topic by estimating the posterior probability $q(z|d)$ of each document from \textit{20NewsGroups}.
Here we set the number of topics to 400, which is large enough for this dataset.
Figure \ref{fig:pmgsm} shows that the topics with higher probability are evenly distributed.
While in Figure \ref{fig:pmgsm} the higher probability ones are placed in the front, and we can see a small tail on the topics after 300.
Due to the sparsity inducing property of the stick-breaking construction, the topics on the tail are less likely to be sampled.
This is also the advantage of stick-breaking construction when we apply the RSB-TF as a non-parameteric topic model, since the model activates the topics according to the knowledge learned from data and it becomes less sensitive to the hyperparameter controlling the initial number of topics.
Figure \ref{fig:np} shows the impact on test perplexity for the neural topic models when the maximum number of topics is increased.
We can see that the performance of the GSM model gets worse if the maximum number of topics exceeds 400, but the GSB and RSB are stable even though the number of topics far outstrips that which the model requires.
In addition, the RSB model performs better than GSB when the number of topics is under 200, but it becomes slightly worse than GSB when the number exceeds 400, possibly due to the difficulty of learning long sequences with RNNs.

\begin{figure}[t]
  \centering
	\includegraphics[width=3.0in]{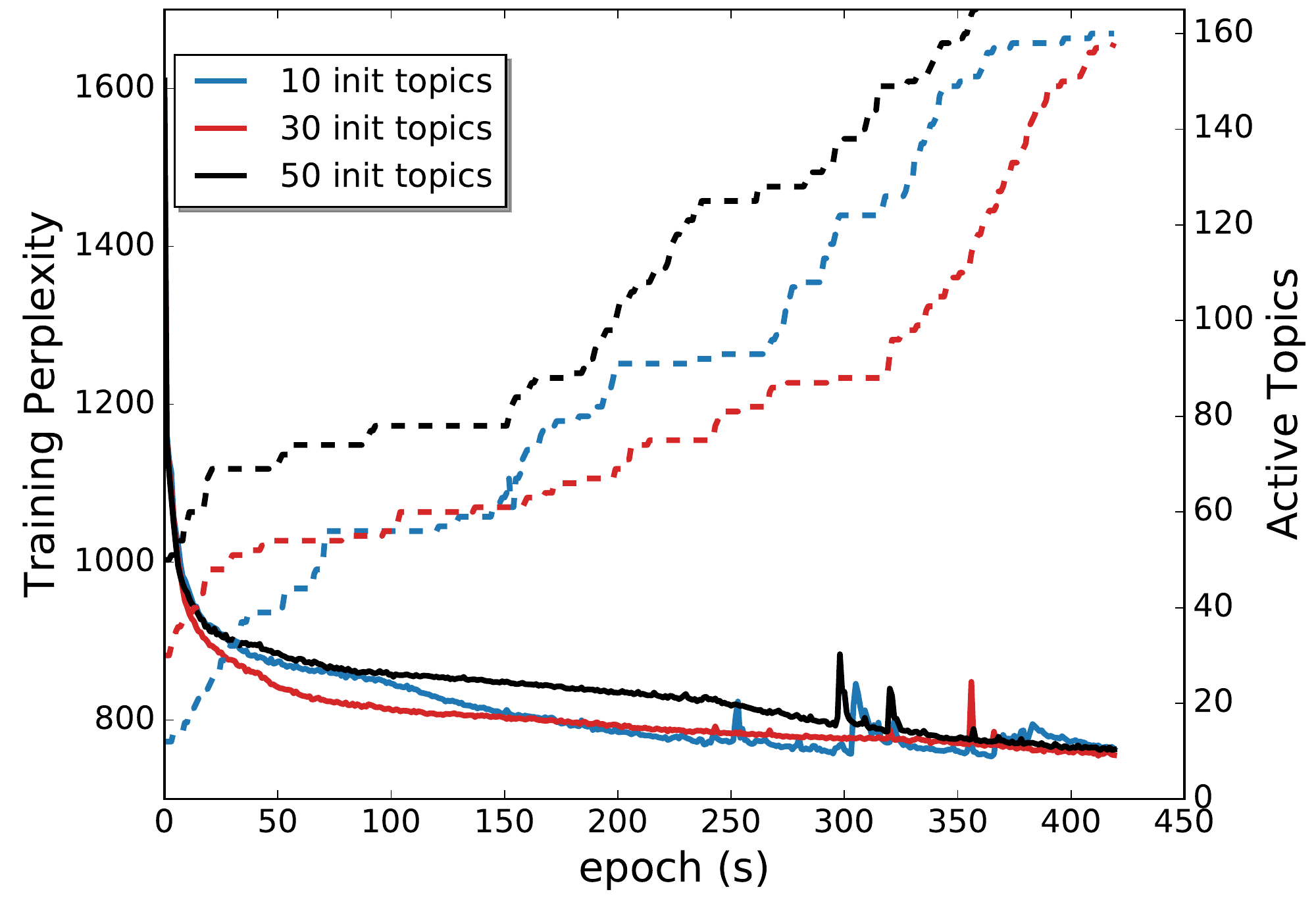}
	\vspace{-1.5em}
  \caption{The convergence behavior of the truncation-free RSB model (RSB-TF) with different initial active topics on \textit{20NewsGroups}. Dash lines represent the corresponding active topics.}
  \label{fig:con5}
  	\vspace{-1em}
\end{figure}

Figure \ref{fig:con5} shows the convergence process of the truncation-free RSB (RSB-TF) model on the \textit{20NewsGroups}. With different initial number of topics, 10, 30, and 50. The RSB-TF dynamically increases the number of active topics to achieve a better variational lower bound.
We can see the training perplexity keeps decreasing while the RSB-TF activates more topics.
The numbers of active topics will stabilise when the convergence point is approaching (normally between 200 and 300 active topics on the \textit{20NewsGroups}).
Hence, as a non-parametric model, RSB-TF is not sensitive to the initial number of active topics.

\begin{figure}[t]
  \centering
  \vspace{-1em}
	\includegraphics[width=2.5in]{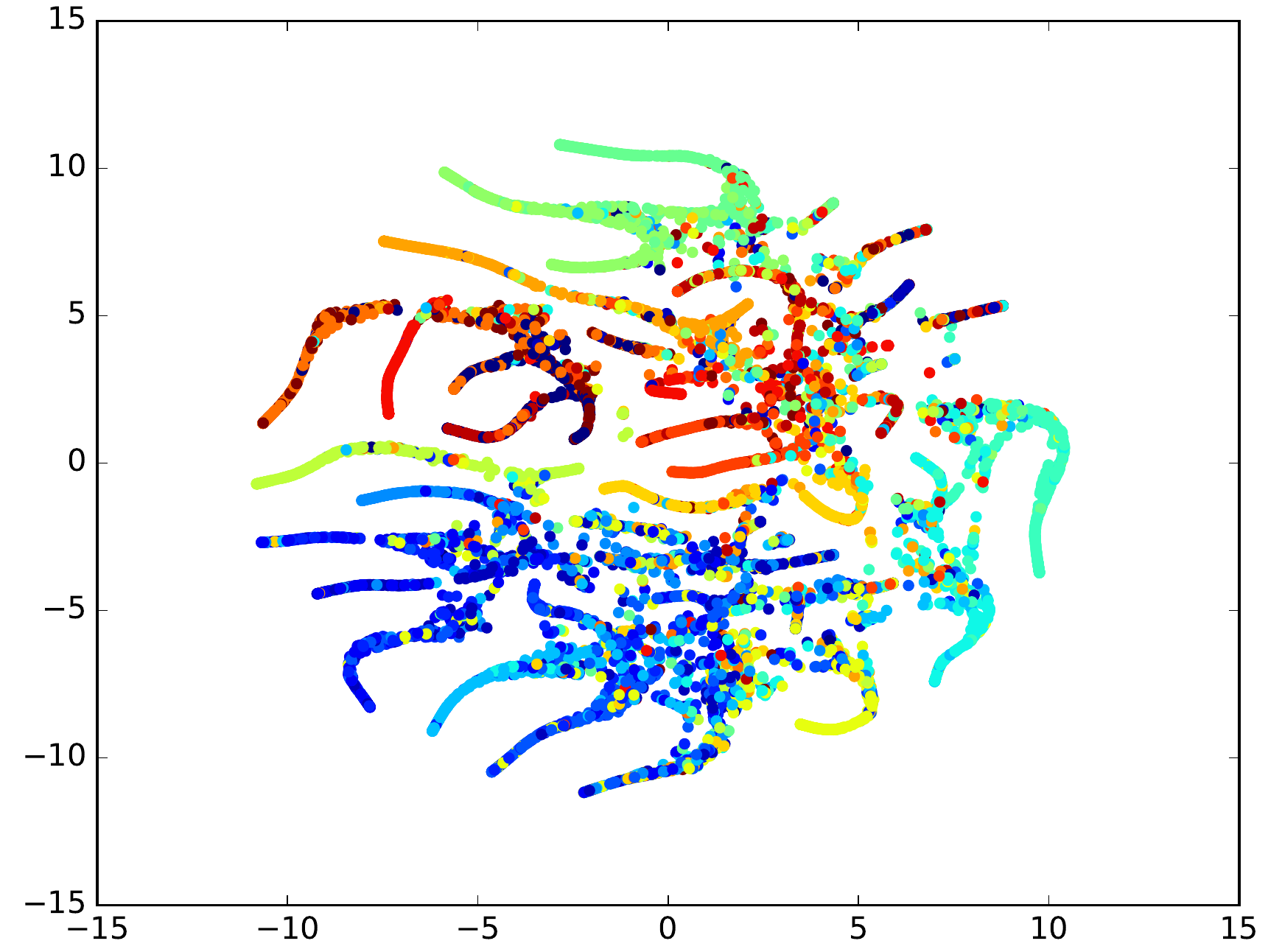}
	\vspace{-0.5em}
  \caption{t-SNE projection of the estimated topic proportions of each document (i.e.~$q(\theta|d)$) from \textit{20NewsGroups}. The vectors are learned by the GSM model with 50 topics and each color represents one group from the 20 different groups of the dataset.}
  \label{fig:tsne}
  	\vspace{-0.5em}
\end{figure}

In addition since the quality of the discovered topics is not directly reflected by perplexity (i.e.~a function of log-likelihood), we evaluate the \textbf{topic observed coherence} by normalised point-wise mutual information (NPMI) \cite{lau2014machine}.
Table \ref{tb:tcntm} shows the topic observed coherence achieved by the finite neural topic models.
According to these results, there does not appear to be a significant difference in topic coherence amongst the neural topic models.
We observe that in both the GSB and RSB, the NPMI scores of the former topics in the stick breaking order are higher than the latter ones.
It is plausible as the stick-breaking construction implicitly assumes the order of the topics, the former topics obtain more sufficient gradients to update the topic distributions.
Likewise we present the results obtained by the neural document models with implicit topic distributions.
Though the topic probability distribution over words does not exist, we could rank the words by the positiveness of the connections between  the words and each dimension of the latent variable.
Interestingly the performance of these document models are significantly better than their topic model counterparts on topic coherence.
The results of RSB-TF and HDP are not presented due to the fact that the number of active topics is dynamic, which makes these two models not directly comparable to the others.
To further demonstrate the quality of the topics, we produce a t-SNE projection for the estimated topic proportions of each document in Figure \ref{fig:tsne}.

\begin{table}[t]
\center
\def\arraystretch{0.8}
\begin{tabular}{lcc}
\toprule[1.2pt]
	\multirow{2}{2cm}{\textbf{Topic Model}} & \multicolumn{2}{c}{Topics} \\
	\cmidrule{2-3}
	 &	50 &	200 	\\
	\hline \\[-1.5ex]
	\hspace{0.2cm}GSM &	0.121 &	0.110 \\
	\hspace{0.2cm}GSB &	0.095 &	0.081	 \\
	\hspace{0.2cm}RSB &	0.111 &	0.097	\\
	\hline \\[-1.5ex]
	\hspace{0.2cm}OnlineLDA & 0.131 & 0.112	\\
	\hspace{0.2cm}NVLDA & 0.110 & 0.110	\\
	\midrule[1.2pt]
	 \multirow{2}{3cm}{\textbf{Document Model}} & \multicolumn{2}{c}{Latent Dimension} \\
	  \cmidrule{2-3} &	50 &	200 	\\
	\hline \\[-1.5ex]
	\hspace{0.2cm}GSM &	0.223 &	0.186	 \\
	\hspace{0.2cm}GSB &	0.217 &	0.171	 \\
	\hspace{0.2cm}RSB &	0.224 &	0.177	\\
	\hline \\[-1.5ex]
	\hspace{0.2cm}NVDM &	0.186 &	0.157  \\
	\hspace{0.2cm}ProdLDA &	0.240 & 	0.190 \tablefootnote{The best scores we obtained are 0.222 and 0.175 for 50 and 200 topics respectively, but here we report the higher scores from \citet{srivastavaneural}.}	\\
\bottomrule[1.2pt]
\end{tabular}
\caption{Topic coherence on \textit{20NewsGroups} (higher is better). We compute coherence over the top-5 words and top-10 words for all topics and then take the mean of both values. }
\label{tb:tcntm}
\end{table}

\section{Conclusion}
In this paper we have introduced a family of neural topic models using the Gaussian Softmax, Gaussian Stick-Breaking and Recurrent Stick-Breaking constructions for parameterising the latent multinomial topic distributions of each document.
With the help of the stick-breaking construction, we are able to build neural topic models which exhibit similar sparse topic distributions as found with traditional Dirichlet-Multinomial models.
By exploiting the ability of recurrent neural networks to model sequences of unbounded length, we further present a truncation-free variational inference method that allows the number of topics to dynamically increase.
The evaluation results show that our neural models achieve state-of-the-art performance on a range of standard document corpora.

\newpage

\bibliography{icml2017}

\begin{thebibliography}{34}
\providecommand{\natexlab}[1]{#1}
\providecommand{\url}[1]{\texttt{#1}}
\expandafter\ifx\csname urlstyle\endcsname\relax
  \providecommand{\doi}[1]{doi: #1}\else
  \providecommand{\doi}{doi: \begingroup \urlstyle{rm}\Url}\fi

\bibitem[Attias(2000)]{attias2000variational}
Attias, Hagai.
\newblock A variational bayesian framework for graphical models.
\newblock In \emph{Proceedings of NIPS}, 2000.

\bibitem[Beal(2003)]{beal2003variational}
Beal, Matthew~James.
\newblock \emph{Variational algorithms for approximate Bayesian inference}.
\newblock University of London, 2003.

\bibitem[Bertin-Mahieux et~al.(2011)Bertin-Mahieux, Ellis, Whitman, and
  Lamere]{Bertin-Mahieux2011}
Bertin-Mahieux, Thierry, Ellis, Daniel~P.W., Whitman, Brian, and Lamere, Paul.
\newblock The million song dataset.
\newblock In \emph{{Proceedings of the 12th International Conference on Music
  Information Retrieval (ISMIR)}}, 2011.

\bibitem[Blei \& Lafferty(2006)Blei and Lafferty]{blei2006dynamic}
Blei, David~M and Lafferty, John~D.
\newblock Dynamic topic models.
\newblock In \emph{Proceedings of ICML}, pp.\  113--120. ACM, 2006.

\bibitem[Blei \& Lafferty(2007)Blei and Lafferty]{blei2007correlated}
Blei, David~M and Lafferty, John~D.
\newblock A correlated topic model of science.
\newblock \emph{The Annals of Applied Statistics}, 2007.

\bibitem[Blei et~al.(2003)Blei, Ng, and Jordan]{blei2003latent}
Blei, David~M, Ng, Andrew~Y, and Jordan, Michael~I.
\newblock Latent dirichlet allocation.
\newblock \emph{The Journal of Machine Learning Research}, 3:\penalty0
  993--1022, 2003.

\bibitem[Bryant \& Sudderth(2012)Bryant and Sudderth]{bryant2012truly}
Bryant, Michael and Sudderth, Erik~B.
\newblock Truly nonparametric online variational inference for hierarchical
  dirichlet processes.
\newblock In \emph{Proceedings of NIPS}, 2012.

\bibitem[Carlin \& Polson(1991)Carlin and Polson]{carlin1991inference}
Carlin, Bradley~P and Polson, Nicholas~G.
\newblock Inference for nonconjugate bayesian models using the gibbs sampler.
\newblock \emph{Canadian Journal of statistics}, 19\penalty0 (4):\penalty0
  399--405, 1991.

\bibitem[Dieng et~al.(2016)Dieng, Wang, Gao, and Paisley]{dieng2016topicrnn}
Dieng, Adji~B, Wang, Chong, Gao, Jianfeng, and Paisley, John.
\newblock Topicrnn: A recurrent neural network with long-range semantic
  dependency.
\newblock \emph{arXiv preprint arXiv:1611.01702}, 2016.

\bibitem[Hinton \& Salakhutdinov(2009)Hinton and
  Salakhutdinov]{hinton2009replicated}
Hinton, Geoffrey~E and Salakhutdinov, Ruslan.
\newblock Replicated softmax: an undirected topic model.
\newblock In \emph{Proceedings of NIPS}, 2009.

\bibitem[Hoffman et~al.(2010)Hoffman, Bach, and Blei]{hoffman2010online}
Hoffman, Matthew, Bach, Francis~R, and Blei, David~M.
\newblock Online learning for latent dirichlet allocation.
\newblock In \emph{Proceedings of NIPS}, pp.\  856--864, 2010.

\bibitem[Hofmann(1999)]{hofmann1999probabilistic}
Hofmann, Thomas.
\newblock Probabilistic latent semantic indexing.
\newblock In \emph{Proceedings of SIGIR}, 1999.

\bibitem[Jordan et~al.(1999)Jordan, Ghahramani, Jaakkola, and
  Saul]{jordan1999introduction}
Jordan, Michael~I, Ghahramani, Zoubin, Jaakkola, Tommi~S, and Saul, Lawrence~K.
\newblock An introduction to variational methods for graphical models.
\newblock \emph{Machine learning}, 37\penalty0 (2):\penalty0 183--233, 1999.

\bibitem[Khan et~al.(2012)Khan, Mohamed, Marlin, and Murphy]{khan2012stick}
Khan, Mohammad~Emtiyaz, Mohamed, Shakir, Marlin, Benjamin~M, and Murphy,
  Kevin~P.
\newblock A stick-breaking likelihood for categorical data analysis with latent
  gaussian models.
\newblock In \emph{Proceedings of AISTATS}, 2012.

\bibitem[Kingma \& Ba(2015)Kingma and Ba]{DBLP:journals/corr/KingmaB14}
Kingma, Diederik~P. and Ba, Jimmy.
\newblock Adam: {A} method for stochastic optimization.
\newblock In \emph{Proceedings of ICLR}, 2015.

\bibitem[Kingma \& Welling(2014)Kingma and Welling]{kingma2013auto}
Kingma, Diederik~P and Welling, Max.
\newblock Auto-encoding variational bayes.
\newblock In \emph{Proceedings of ICLR}, 2014.

\bibitem[Landauer et~al.(1998)Landauer, Foltz, and
  Laham]{landauer1998introduction}
Landauer, Thomas~K, Foltz, Peter~W, and Laham, Darrell.
\newblock An introduction to latent semantic analysis.
\newblock \emph{Discourse processes}, 25\penalty0 (2-3):\penalty0 259--284,
  1998.

\bibitem[Larochelle \& Lauly(2012)Larochelle and Lauly]{larochelle2012neural}
Larochelle, Hugo and Lauly, Stanislas.
\newblock A neural autoregressive topic model.
\newblock In \emph{Proceedings of NIPS}, 2012.

\bibitem[Lau et~al.(2014)Lau, Newman, and Baldwin]{lau2014machine}
Lau, Jey~Han, Newman, David, and Baldwin, Timothy.
\newblock Machine reading tea leaves: Automatically evaluating topic coherence
  and topic model quality.
\newblock In \emph{Proceedings of EACL}, pp.\  530--539, 2014.

\bibitem[LeCun et~al.(2006)LeCun, Chopra, and Hadsell]{lecun2006tutorial}
LeCun, Yann, Chopra, Sumit, and Hadsell, Raia.
\newblock A tutorial on energy-based learning.
\newblock \emph{Predicting structured data}, 2006.

\bibitem[Mcauliffe \& Blei(2008)Mcauliffe and Blei]{mcauliffe2008supervised}
Mcauliffe, Jon~D and Blei, David~M.
\newblock Supervised topic models.
\newblock In \emph{Advances in neural information processing systems}, pp.\
  121--128, 2008.

\bibitem[Miao et~al.(2016)Miao, Yu, and Blunsom]{miao2015neural}
Miao, Yishu, Yu, Lei, and Blunsom, Phil.
\newblock Neural variational inference for text processing.
\newblock In \emph{Proceedings of ICML}, 2016.

\bibitem[Mnih \& Gregor(2014)Mnih and Gregor]{mnih2014neural}
Mnih, Andriy and Gregor, Karol.
\newblock Neural variational inference and learning in belief networks.
\newblock In \emph{Proceedings of ICML}, 2014.

\bibitem[Mnih et~al.(2014)Mnih, Heess, and Graves]{mnih2014recurrent}
Mnih, Volodymyr, Heess, Nicolas, and Graves, Alex.
\newblock Recurrent models of visual attention.
\newblock In \emph{Proceedings of NIPS}, 2014.

\bibitem[Nalisnick \& Smyth(2016)Nalisnick and Smyth]{nalisnick2016deep}
Nalisnick, Eric and Smyth, Padhraic.
\newblock Deep generative models with stick-breaking priors.
\newblock \emph{arXiv preprint arXiv:1605.06197}, 2016.

\bibitem[Rezende et~al.(2014)Rezende, Mohamed, and
  Wierstra]{rezende2014stochastic}
Rezende, Danilo~J, Mohamed, Shakir, and Wierstra, Daan.
\newblock Stochastic backpropagation and approximate inference in deep
  generative models.
\newblock In \emph{Proceedings of ICML}, 2014.

\bibitem[Rosen-Zvi et~al.(2004)Rosen-Zvi, Griffiths, Steyvers, and
  Smyth]{rosen2004author}
Rosen-Zvi, Michal, Griffiths, Thomas, Steyvers, Mark, and Smyth, Padhraic.
\newblock The author-topic model for authors and documents.
\newblock In \emph{Proceedings of the 20th conference on Uncertainty in
  artificial intelligence}, pp.\  487--494. AUAI Press, 2004.

\bibitem[Sethuraman(1994)]{sethuraman1994constructive}
Sethuraman, Jayaram.
\newblock A constructive definition of dirichlet priors.
\newblock \emph{Statistica sinica}, pp.\  639--650, 1994.

\bibitem[Srivastava \& Sutton(2016)Srivastava and Sutton]{srivastavaneural}
Srivastava, Akash and Sutton, Charles.
\newblock Neural variational inference for topic models.
\newblock \emph{Bayesian deep learning workshop, NIPS 2016}, 2016.

\bibitem[Teh et~al.(2006)Teh, Jordan, Beal, and Blei]{teh2006hierarchical}
Teh, Yee~Whye, Jordan, Michael~I, Beal, Matthew~J, and Blei, David~M.
\newblock Hierarchical dirichlet processes.
\newblock \emph{Journal of the American Statistical Asociation}, 101\penalty0
  (476), 2006.

\bibitem[Wang \& Blei(2013)Wang and Blei]{wang2013variational}
Wang, Chong and Blei, David~M.
\newblock Variational inference in nonconjugate models.
\newblock \emph{Journal of Machine Learning Research}, 14\penalty0
  (Apr):\penalty0 1005--1031, 2013.

\bibitem[Wang et~al.(2011)Wang, Paisley, and Blei]{wang2011online}
Wang, Chong, Paisley, John~William, and Blei, David~M.
\newblock Online variational inference for the hierarchical dirichlet process.
\newblock In \emph{Proceedings of AISTATS}, 2011.

\bibitem[Wang \& McCallum(2006)Wang and McCallum]{wang2006topics}
Wang, Xuerui and McCallum, Andrew.
\newblock Topics over time: a non-markov continuous-time model of topical
  trends.
\newblock In \emph{Proceedings of the 12th ACM SIGKDD international conference
  on Knowledge discovery and data mining}, pp.\  424--433. ACM, 2006.

\bibitem[Xie et~al.(2015)Xie, Deng, and Xing]{xie2015diversifying}
Xie, Pengtao, Deng, Yuntian, and Xing, Eric.
\newblock Diversifying restricted boltzmann machine for document modeling.
\newblock In \emph{Proceedings of KDD}, pp.\  1315--1324. ACM, 2015.

\end{thebibliography}
\bibliographystyle{icml2017}

\newpage

\quad

\newpage
\begin{appendices}

\section{Discovered Topics}

Table \ref{tb:topics2} presents the topics by the words with highest probability (top-10 words) achieved by different neural topic models on \textit{20NewsGroups} dataset.
\begin{table}[H]
\subfloat[Topics learned by GSM.]{
\addtolength{\tabcolsep}{-3.5pt}
  \small
  \centering
\begin{tabular}{c|c|c|c|c}
\toprule[1.2pt]
	\textit{\textbf{Space}}&		\textit{\textbf{Religion}}&	\textit{\textbf{Encryption}}&		\textit{\textbf{Sport}}&		\textit{\textbf{Science}}	\\
\hline
space & god & encryption & player & science \\
satellite & atheism & device & hall & theory \\
april & exist & technology & defensive & scientific \\
sequence & atheist & protect & team & universe \\
launch & moral & americans & average & experiment \\
president & existence & chip & career & observation \\
station & marriage & use & league & evidence \\
radar & system & privacy & play & exist \\
training & parent & industry & bob & god \\
committee & murder & enforcement & year & mistake \\
\bottomrule[1.2pt]
\end{tabular}
\label{tb:nw1}
}
\hfill
\subfloat[Topics learned by GSB.]{
\addtolength{\tabcolsep}{-3.5pt}
  \small
  \centering
\begin{tabular}{c|c|c|c|c}
\toprule[1.2pt]
	\textit{\textbf{Space}}&		\textit{\textbf{Religion}}&	\textit{\textbf{Lawsuit}}&		\textit{\textbf{Vehicle}}&		\textit{\textbf{Science}}	\\
\hline
moon & atheist & homicide & bike & theory \\
lunar & life & gun & motorcycle & science \\
orbit & eternal & rate & dod & gary \\
spacecraft & christianity & handgun & insurance & scientific \\
billion & hell & crime & bmw & sun \\
launch & god & firearm & ride & orbit \\
space & christian & weapon & dealer & energy \\
hockey & atheism & knife & oo & experiment \\
cost & religion & study & car & mechanism \\
nasa & brian & death & buy & star \\
\bottomrule[1.2pt]
\end{tabular}
\label{tb:nw2}
}
\hfill
\subfloat[Topics learned by RSB.]{
\addtolength{\tabcolsep}{-3.5pt}
  \small
  \centering
\begin{tabular}{c|c|c|c|c}
\toprule[1.2pt]
	\textit{\textbf{Aerospace}}&		\textit{\textbf{Crime}}&	\textit{\textbf{Hardware}}&		\textit{\textbf{Technology}}&		\textit{\textbf{Science}}	\\
\hline
instruction & gun & drive & technology & science \\
spacecraft & weapon & scsi & americans & hell \\
amp & crime & ide & pit & scientific \\
pat & firearm & scsus & encryption & evidence \\
wing & criminal & hd & policy & physical \\
plane & use & go & industry & eternal \\
algorithm & control & controller & protect & universe \\
db & handgun & tape & privacy & experiment \\
reduce & law & datum & product & reason \\
orbit & kill & isa & approach & death \\
\bottomrule[1.2pt]
\end{tabular}
\label{tb:nw3}
}
\caption{Topics learned by neural topic models on 20NewsGroups dataset.}
\label{tb:topics2}
\end{table}

\newpage

\section{Topic Diversity}
\label{a:td}

An issue that exists in both probabilistic and neural topic models is redundant topics.
In neural models, however, we are able to straightforwardly regularise the distance between each of the topic vectors in order to diversify the topics.
Following \citet{xie2015diversifying}, we apply such topic diversity regularisation during the inference process.
We compute the angles between each two topics $ a(t_i,t_j) = arccos(\frac{ | t_i  \cdot t_j |}{ ||t_i || \cdot ||t_j || }) $.
Then, the mean angle of all pairs of $K$ topics is $\zeta=\frac{1}{K^2} \sum_i\sum_j a(t_i,t_j)$, and the variance is $\nu=\frac{1}{K^2} \sum_i\sum_j (a(t_i,t_j)-\zeta)^2$.
We add the following topic diversity regularisation to the variational objective:
\begin{eqnarray*}
   \mathcal{J} = \mathcal{L} + \lambda (\zeta - \nu) ,
\end{eqnarray*}
where $\lambda$ is a hyper-parameter for the regularisation that is set as 0.1 in the experiments. 
During training, the mean angle is encouraged to be larger while the variance is suppressed to be smaller so that all of the topics will be pushed away from each other in the topic semantic space. Though in practice diversity regularisation does not provide a significant improvement to perplexity ($2\!\!\sim\!\! 5$ in most cases), it helps reduce topic redundancy and can be easily applied on topic vectors instead of the simplex over the full vocabulary.

\end{appendices}

\end{document}